\documentclass[eng,article,submit,pdftex,moreauthors]{Definitions/mdpi} 
\firstpage{1}
\pubvolume{1}
\issuenum{1}
\articlenumber{0}
\pubyear{2022}
\copyrightyear{2022}
\datereceived{}
\dateaccepted{}
\datepublished{}
\hreflink{https://doi.org/}

\usepackage{booktabs}
\usepackage{subcaption}

\graphicspath{{Figures/}}

\Title{EVALUATION OF COLOR ANOMALY DETECTION IN MULTISPECTRAL IMAGES FOR SYNTHETIC APERTURE SENSING}

\TitleCitation{Evaluation of Color Anomaly Detection in Multispectral Images for Synthetic Aperture Sensing}

\Author{Francis Seits, Indrajit Kurmi and Oliver Bimber *}

\AuthorNames{Francis Seits, Indrajit Kurmi and Oliver Bimber}

\AuthorCitation{Seits, F.; Kurmi, I.; Bimber, O.}

\address{Institute of Computer Graphics, Johannes Kepler University Linz, 4040 Linz, Austria
\corres{Correspondence: oliver.bimber@jku.at; Tel.: +43-732-2468-6631}}

\abstract{
	In this article we evaluate unsupervised anomaly detection methods in multispectral images obtained with a wavelength-independent synthetic aperture sensing technique, called  Airborne Optical Sectioning (AOS).
	With a focus on search and rescue missions that apply drones to locate missing or injured persons in dense forest and require real-time operation, we evaluate runtime vs. quality of these methods. Furthermore, we show that color anomaly detection methods that normally operate in the visual range always benefit from an additional far infrared (thermal) channel. We also show that, even without additional thermal bands, the choice of color space in the visual range already has an impact on the detection results. Color spaces like HSV and HLS have the potential to outperform the widely used RGB color space, especially when color anomaly detection is used for forest-like environments.
}

\keyword{
	Multispectral, Image Processing, Anomaly Detection, Search and Rescue, Unmanned Aerial Vehicles, Airborne Optical Sectioning.
} 

\begin{document}

\section{Introduction}\label{introduction}
Color anomaly detection methods identify pixel regions in multispectral images that have a low probability of occurring in the background landscape, and are therefore considered to be outliers. Such techniques are used in remote sensing applications for agriculture, wildlife observation, surveillance, or search and rescue. Occlusion caused by vegetation, however, remains a major challenge.     

\begin{figure}[H]
	\centering
	\includegraphics[width=\linewidth]{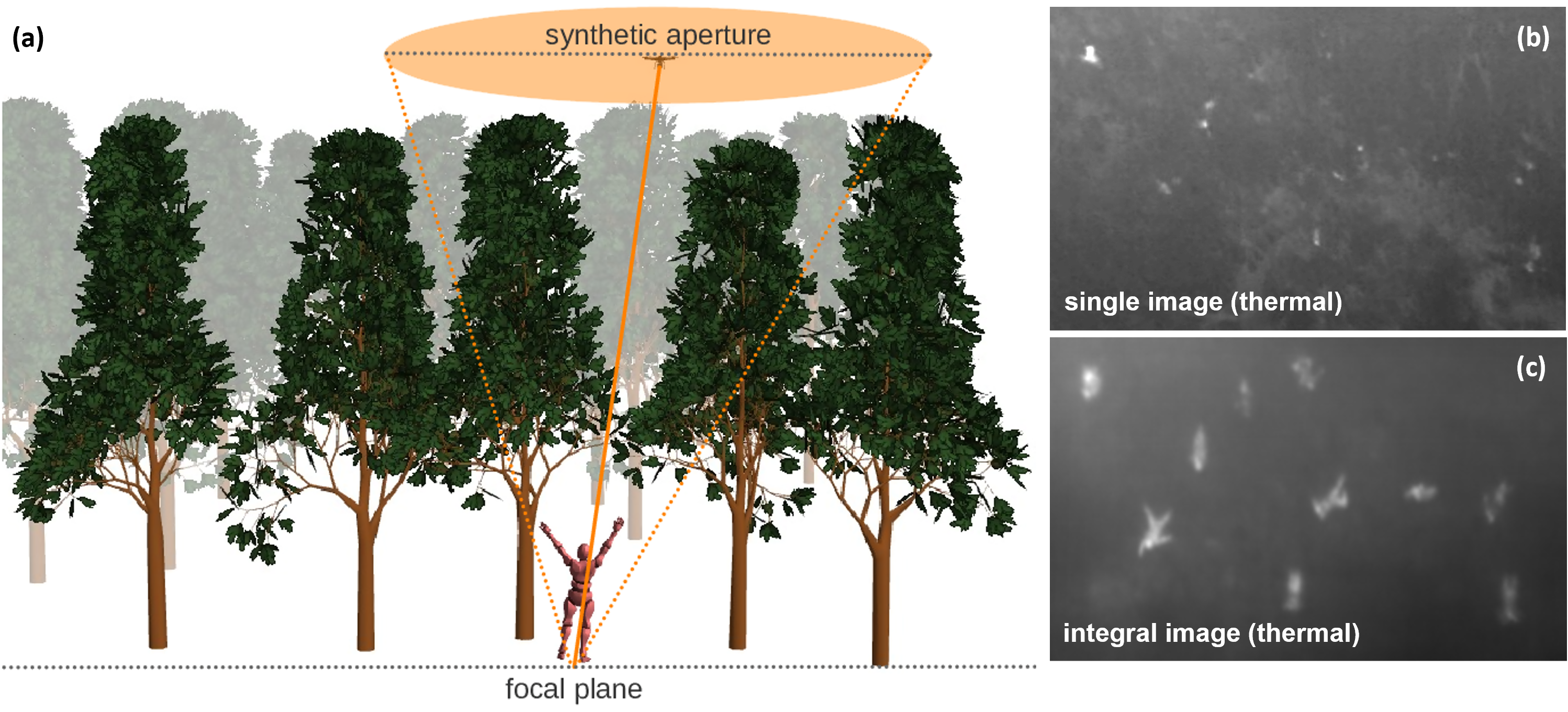}
	\caption{
		Airborne Optical Sectioning (AOS) is a synthetic aperture sensing technique that computationally combines multiple aerial images captured within a synthetic aperture area (a) to an integral image which enhances targets on the synthetic focal plane while suppressing occluders above it. Right: People covered by forest canopy. Single aerial image (thermal channel) that suffers from strong occlusion (b), and corresponding integral image of the same environment with occlusion removed (c).    
	}
	\label{aos}
\end{figure}

Airborne Optical Sectioning (AOS) \cite{kurmi2018a,bimber2019a,kurmi2019a,kurmi2019b,schedl2020a,kurmi2021a,kurmi2021b,schedl2020b,schedl2021a,kurmi2022a,nathan2022a,seits2022a,nathan2022b} is a synthetic aperture sensing technique that computationally removes occlusion in real-time by registering and integrating multiple images captured within a large synthetic aperture area above forest (cf. Fig. \ref{aos}). With  resulting, shallow-depth-of-field integral images, it becomes possible to locate targets (e.g., people, animals, vehicles, wildfires, etc.) that are otherwise hidden under the forest canopy. Image pixels that correspond to the same target on the synthetic focal plane (i.e., the forest ground) are computationally aligned and enhanced while occluders above the focal plane (i.e., trees) are suppressed in strong defocus. AOS is real-time and wavelength independent (i.e., it can be applied to images in all spectral bands), which is beneficial for many areas of application. 

Anomaly detection methods for wilderness search and rescue have been evaluated earlier \cite{morse2012a}, and bimodal systems using a composition of visible and thermal information were already used to improve detection rates of machine learning algorithms \cite{rudol2008a,hinzmann2020a}. However, none of the previous work considered occlusion. 

With AOS we are able to combine multispectral recordings into a single integral image. Our previous work has shown that image processing tasks, like person classification with deep neural networks \cite{schedl2020b,schedl2021a,kurmi2022a} perform significantly better on integral images when compared to single images. These classifiers are based on supervised architectures, which have the disadvantage that training data must be collected and labeled in a time-consuming manner and that the trained neural networks do not generalize well into other domains. It was also shown in \cite{nathan2022a} that the image integration process of AOS decreases variance and co-variance, which allows better separation of target and background pixels when applying Reed-Xiaoli (RX) unsupervised anomaly detection \cite{reed1990a}.

In this article, we evaluate several common unsupervised anomaly detection methods being applied to multispectral integral images that are captured from a drone when flying over open and over occluded (forest) landscapes. We show that their performance can significantly be improved by the right combination of spectral bands and choice of color space input format. Especially for forest-like environments, detection rates of occluded people can be consistently increased if visible and thermal bands are combined, and if HSV or HLS color spaces are used for the visible bands instead of common RGB. Furthermore, we also evaluate runtime behaviour of these methods when considered for time-critical applications, such as search and rescue.

\section{Materials and Methods}\label{methods}
\begin{figure}[H]
	\begin{subfigure}{0.5\linewidth}
		\centering
		\includegraphics[width=0.95\linewidth]{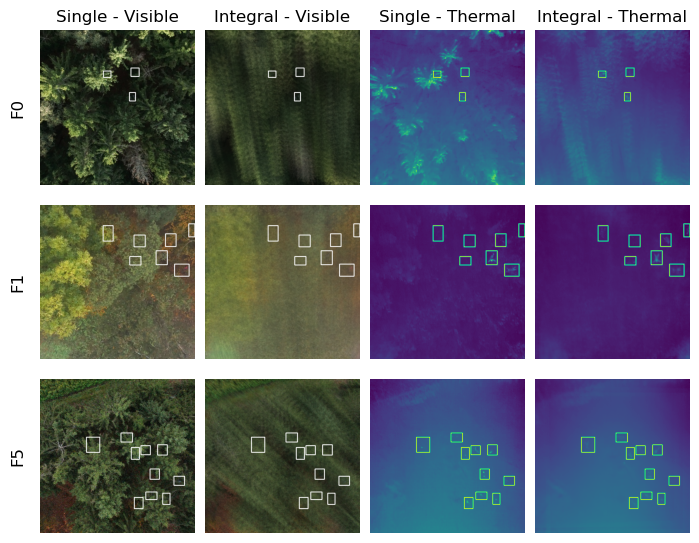}
		\caption{Forest Landscapes}
		\label{scenes.forest}
	\end{subfigure}
	\begin{subfigure}{0.5\linewidth}
		\centering
		\includegraphics[width=0.95\linewidth]{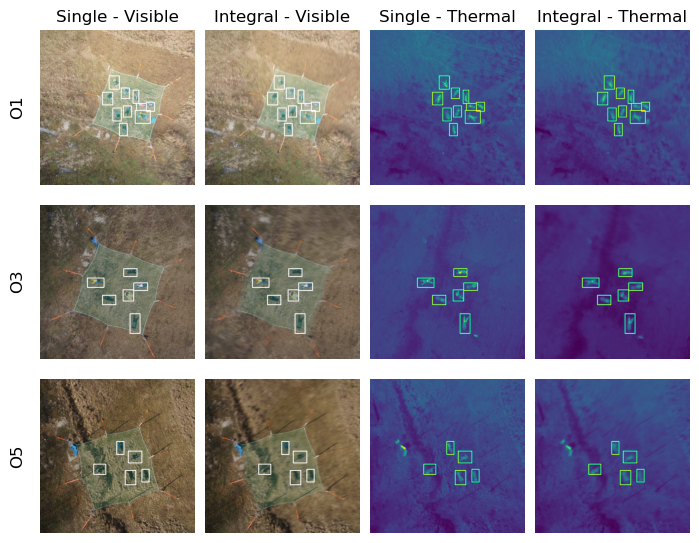}
		\caption{Open Landscapes}
		\label{scenes.open}
	\end{subfigure}
	\caption{
		Our evaluation dataset consists of several forest (a) and open (b) landscape images captured with a drone from an altitude of about 35m AGL. Each scenery (F0,F1,F5,O1,O3,O5) contains about 20 consecutive single images taken in the visible (RGB) and thermal spectrum which are combined to integral images. Rectangles indicate manually labeled persons lying on the ground.
	}
	\label{scenes}
\end{figure}

For our evaluation, we apply the dataset from \cite{schedl2020b}, which was used to prove that  integral images improve people classification under occluded conditions. It consists of RGB and thermal images (pairwise simultaneously) captured with a drone prototype over multiple forests types (broadleaf, conifer, mixed) and open landscapes as shown in Fig. \ref{scenes}. In all images, targets (persons laying on the ground) are manually labeled. Additional telemetry data (GPS and IMU sensor values) of the drone during capturing is also provided for each image.

While the visible bands were converted from RGB to other color spaces (HLS, HSV, LAB, LUV, XYZ, and YUV), the thermal data was optionally added as a fourth (alpha) channel, resulting in additional input options (RGB-T, HLS-T, HSV-T, LAB-T, LUV-T, XYZ-T, and YUV-T).   

All images had a resolution of 512x512 pixels, so the input dimensions where either (512, 512, 3) or (512, 512, 4). Methods that do not require spatial information used flattened images with (262144, 3) or (262144, 4) dimensions.

The publicly available C/C++ implementation of AOS \footnote{Source Code: \url{https://github.com/JKU-ICG/AOS}} was used to compute integral images from single images.

\subsection{Color Anomaly Detectors}\label{detectors}
Unsupervised color anomaly detectors have been widely used in the past \cite{reed1990a,chang2002a,shyu2003a,bishop2007a,carlotto2005a,breunig2000a}, with the Reed-Xiaoli (RX) detector \cite{reed1990a} being commonly considered as a benchmark.
Several variations of RX exist, where the standard implementation calculates global background statistics (over the entire image) and then compares individual pixels based on the Mahalanobis distance.
In the further course of this article we will refer to this particular RX detector as Reed-Xiaoli Global (RXG).

The following summarized briefly the considered color anomaly detectors, while details can be found through the provided references:

\vspace{10pt}
The Reed-Xiaoli Global (RXG) detector \cite{reed1990a} computes a $K_{n \times n}$ covariance matrix of the image, where $n$ is given by the number of input channels (e.g., for RGB, $n=3$ and for RGB-T, $n=4$). The pixel under test is the $n$-dimensional vector $r$ and the mean is given by the $n$-dimensional vector $\mu$:
\begin{equation*}
	\alpha_{RXG}(r) = (r - \mu)^T K_{n \times n}^{-1}(r - \mu).
\end{equation*}

\vspace{10pt}
The Reed Xiaoli Modified (RXM) detector \cite{chang2002a} is a variation of RXG, where an additional constant $\kappa = ||r - \mu||^{-1}$ is used for normalization:
\begin{equation*}
	\alpha_{RXM}(r) = \kappa \cdot \alpha_{RXG}(r) = \left(\frac{r - \mu}{||r - \mu||}\right)^T K_{n \times n}^{-1}(r - \mu).
\end{equation*}

\vspace{10pt}
The Reed Xiaoli Local (RXL) detector computes covariance and mean over smaller local areas and therefore does not use global background statistics.
The areas are defined by an inner window ($guard\_win$) and an outer window ($bg\_win$).
The mean $\mu$ and covariance $K$ are calculated based on the outer window, but excludes the inner window.
Window sizes where chosen to be $guard\_win=33$ and $bg\_win=55$, based on the projected pixel sizes of the targets in the forest landscape.

\vspace{10pt}
The Principal Component Analysis (PCA) \cite{shyu2003a} uses singular value decomposition for a linear dimensionality reduction.
The covariance matrix of the image is decomposed into eigenvectors and there corresponding eigenvalues.
A low dimensional hyperplane is constructed by selected ($n\_components$) eigenvectors.
Outlier scores for each samples are then obtained by their euclidean distance to the constructed hyperplane. The number of eigenvectors to use was chosen to be $n\_components=n$, where $n$ is given by the number of input channels

\vspace{10pt}
The Gaussian Mixture Model (GMM) \cite{bishop2007a} is a clustering approach, where multiple Gaussian distributions are used to characterize the data.
The data is fit to each of the single Gaussians ($n\_components$), which are considered as a representation of clusters.
For each sample the algorithm calculates the probability of belonging to each cluster, where low probabilities are an indication of being an anomaly.
The number of Gaussians to use was chosen to be $n\_components=2$.

\vspace{10pt}
The Cluster Based Anomaly Detection (CBAD) \cite{carlotto2005a} estimates background statistics over clusters, instead of sliding windows.
The image background is partitioned (using any clustering algorithm) into clusters ($n\_cluster$), where each cluster can be modeled as a Gaussian distribution.
Similar to GMM, anomalies have values that deviate significantly from the cluster distributions.
Samples are each assigned to the nearest background cluster, becoming an anomaly if their value deviates farther from the mean than background pixel values in that cluster.
The number of clusters to use was chosen to be $n\_cluster=2$.

\vspace{10pt}
The Local Outlier Factor (LOF) \cite{breunig2000a} uses a distance metric (e.g., Minkowski distance) to determine the distances between neighboring ($n\_neighbors$) data points.
Based on the inverse of those average distances, the local density is calculated.
This is then compared to the local densities of there surrounding neighborhood.
Samples that have significantly lower densities than their neighbors are considered isolated and are therefore becoming an outlier.
The number of neighbors to use was chosen to be $n\_neighbors=200$.

\subsection{Evaluation}\label{evaluation}
The evaluation of the methods summarized above was carried out on a consumer PC (Intel Core i9-11900H @ 2.50GHz) for the landscapes shown in Fig. \ref{scenes}.

Precision (Eqn. \ref{precision}) vs. recall (Eqn. \ref{recall}) was used as metrics for performance comparisons.

The task can be formulated as a binary classification problem, where positive predictions are considered as anomalously pixels. The data we want to classify (image pixels) is highly unbalanced, as most of the pixels are considered as background (majority class) and only some of the pixels are considered as anomalies (minority class).

The true positive (TP) pixels are determined by checking whether they lie within one of the labeled rectangles, as shown in Fig. \ref{scenes}. Pixels detected outside these rectangles are considered as false positives (FP) and pixels inside the rectangle but not classified as anomalously are considered as false negatives (FN).
Since the dataset only provides rectangles for labels and not perfect masks around the persons, the recall results are biased (in general less good as expected). As we are mainly interested in the performance difference between individual methods and the errors introduced are always constant (rectangle area - real person mask), the conclusions drawn from the results should be the same, even if perfect masks were used instead.

The precision (Eqn. \ref{precision}) quantifies the number of correct positive predictions made and recall (Eqn. \ref{recall}) quantifies the number of correct positive predictions made out of all positive predictions that could have been made. 

\begin{equation}\label{precision}
	Precision = \frac{TP}{TP + FP}
\end{equation}
\begin{equation}\label{recall}
	Recall = \frac{TP}{FN + TP}
\end{equation}

Precision and recall both focuses on the minority class (anomalously pixels) and are therefore less concerned with the majority class (background pixels), which is important for our unbalanced dataset. 

Since the anomaly detection methods provide probabilistic scores on the likelihood of a pixel being considered anomalous, a threshold value must be chosen to obtain a final binary result.

\begin{figure}[H]
	\centering
	\includegraphics[width=\linewidth/2]{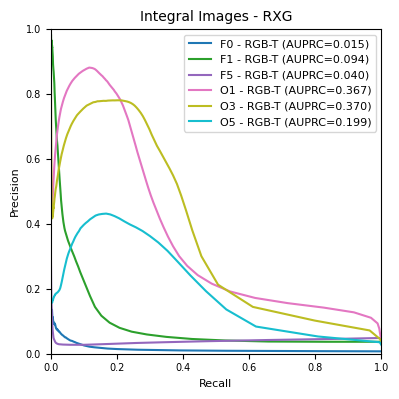}
	\caption{
		The area under the precision-recall curve (AUPRC) is used as metric for comparing the performance of the evaluated anomaly detection methods.
		We consider true positives (TP), false positives (FP), and false negatives (FN) pixels for each image and calculate precision and recall. The example above illustrates precision-recall curves of all landscapes for RXG and with RGB-T input.	
	}
	\label{curve}
\end{figure}

The precision-recall curve (PRC) in Fig. \ref{curve} shows the relationship between precision and recall for every possible threshold value that could be chosen. Thus, a method performing well would have high precision and high recall over different threshold values.
We use the area under the precision-recall curve (AUPRC), which is simply the integral of the PRC, as the final evaluation metric.

The AUPRC metric provides comparable results on the overall performance of a method, but is not well suited when it comes to finding the best threshold for a single image.
To obtain the best threshold value for a single image we use the F$_\beta$-score (Eqn. \ref{fscore}), which is also calculated from precision and recall:

\begin{equation}\label{fscore}
	F_\beta = \left(1 + \beta^2\right) \cdot \frac{Precision \cdot Recall}{\left(\beta^2 \cdot Precision \right) + Recall},
\end{equation}

where $\beta$ is used as a weighting factor that can be chosen such that recall is considered $\beta$-times more important than precision.

The balanced F$_1$-score is the harmonic mean of precision and recall and is widely used. But as we care more about minimizing false positives than minimizing false negatives, we would select a $\beta < 1$.
Thus, with $\beta = \frac{1}{2}$, precision is given more weight than recall.
The $F_\beta$ metric is only used to threshold the image scores for comparison purposes as shown in Fig. \ref{scores}.

\section{Results}\label{results}
\begin{figure}[H]
	\includegraphics[width=1.0\linewidth]{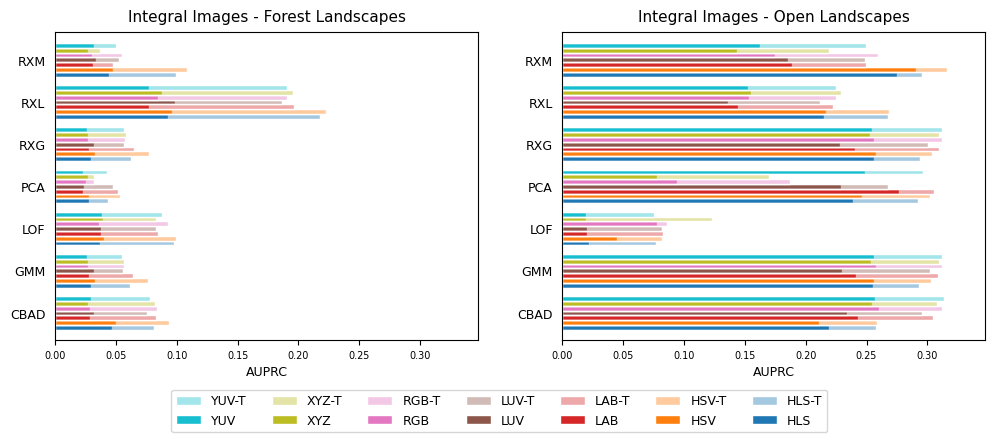}
	\caption{
		Results of area under the precision-recall curve (AUPRC) values for multiple color spaces and color anomaly detection methods. The results of the forest landscape are average values over F0, F1, and F5 and the results of the open landscape are average values over O1, O3, and O5. The stacked bar charts highlight the improvement gains caused by the additional thermal channel.
	}
	\label{area}
\end{figure}

Fig. \ref{area} shows the AUPRC values across different color spaces and methods. The methods are evaluated on each color space, once with three channels (visible spectrum only) and once with four channels (visible and thermal spectrum).
The results of the forest landscape are average values over F0, F1 and F5, and the results of the open landscape are average values over O1, O3 and O5.

As expected (and as we have also seen in Fig. \ref{curve}), the overall AUPRC of the open landscapes is much higher than the AUPRC values of the more challenging forest landscapes. The reason is occlusion in the presence of forests.

The AUPRC values of the four channel (color+thermal) and three channel (color only) inputs are overlayed in the same bar. The slightly lighter colored four channel results are always higher than the three channel results -- regardless of the method or the color space used. Yet, the difference is more pronounced for the forest landscapes than for the open landscapes.
This shows that, regardless of the scenery and regardless of the method and the color spaced used, the additional thermal information always improves the performance of anomaly detection.

With a look at the AUPRC values in the forest landscapes we can observe that RXL gives overall the best results and outperforms all other methods. 
Utilizing the additional thermal information gives in this case even a $2 \times$ gain.
This can also be observed visually in the anomaly detection scores shown in Fig. \ref{scores}, where FP's detections highly decrease and TP's detections highly increase if the thermal channel is added (e.g., F1, in the visible spectral band many background pixels are considered anomalous, with the additional thermal information those misclassified pixels are eliminated).

Looking at the AUPRC values in the open landscapes, we can observe that the difference between the methods is not as pronounced as in the forest landscapes.
An obvious outlier, however, seems to be LOF which nevertheless performs very well (second best) in the forest landscapes.
This can be explained by the fact that hyper-parameters of the methods where specifically chosen for the forest landscape.
In the case of LOF the $n\_neighbors$ parameter was set to be $200$, which seems suboptimal for the open landscapes.
The same holds for RXL (window sizes), CBAD (number of clusters), GMM (number of components) and for PCA (number of components).
All other methods do not require hyper-parametrization.

Another observation that can be made is that some color spaces consistently gives better results than others.
In the forest landscapes, HSV(-T) usually gives the best results, regardless of the methods being used.
In the open landscapes, it is not as clear which color space performs best, but HSV(-T) still gives overall good results. 
In general and especially for RXM, the improvements achieved by choosing HSV(-T) over other color spaces is clearly noticeable.

The individual results plotted in Fig. \ref{area} are also shown in Tab. \ref{auprc.forest} and Tab. \ref{auprc.open}, where the mean values over all color spaces (last row) may give a useful estimate on the methods' overall performance.
Since for time-critical applications, anomaly detection should deliver reliable results in real-time, we have also measured their runtimes, as shown in Tab. \ref{runtime}.
The best performing method on the forest landscapes in terms of AUPRC values are RXL and LOF.
In terms of runtime, both are found to be very slow, as they consume 20 to 35 seconds for computations, where all other algorithms provide anomaly scores in under a second (cf. Fig. \ref{auprc.runtime}).

\begin{table}
	\centering
	\setlength{\tabcolsep}{4.3pt}
	\begin{tabular}{lrrrrrrr}
		\toprule
		{} &  \textbf{CBAD} &  \textbf{GMM} &  \textbf{LOF} &  \textbf{PCA} &  \textbf{RXG} &  \textbf{RXL} &  \textbf{RXM} \\
		\midrule
		\textbf{HLS} &   887 &  219 &  18440 &   98 &   39 &  36647 &   28 \\
		\textbf{HSV} &   883 &  225 &  18094 &  102 &   40 &  36346 &   27 \\
		\textbf{LAB} &   877 &  219 &  19833 &   99 &   36 &  36495 &   29 \\
		\textbf{LUV} &   888 &  212 &  19975 &   96 &   41 &  36166 &   29 \\
		\textbf{RGB} &   925 &  216 &  19367 &   96 &   40 &  35732 &   29 \\
		\textbf{XYZ} &   872 &  224 &  19107 &  100 &   40 &  36367 &   27 \\
		\textbf{YUV} &   874 &  228 &  20485 &  100 &   42 &  36409 &   27 \\
		\midrule
		\text{Runtime} &   887 &  221 &  19329 &   99 &   40 &  36309 &   28 \\
		\bottomrule \\
	\end{tabular}
	\begin{tabular}{lrrrrrrr}
		\toprule
		{} &  \textbf{CBAD} &  \textbf{GMM} &  \textbf{LOF} &  \textbf{PCA} &  \textbf{RXG} &  \textbf{RXL} &  \textbf{RXM} \\
		\midrule
		\textbf{HLS-T} &   878 &  243 &  28054 &  104 &   43 &  36002 &   30 \\
		\textbf{HSV-T} &   882 &  237 &  28235 &  107 &   40 &  35801 &   30 \\
		\textbf{LAB-T} &   879 &  230 &  28005 &  108 &   32 &  35876 &   31 \\
		\textbf{LUV-T} &   893 &  229 &  27506 &  105 &   45 &  36059 &   32 \\
		\textbf{RGB-T} &   904 &  230 &  27323 &   98 &   37 &  35618 &   32 \\
		\textbf{XYZ-T} &   880 &  241 &  24395 &  108 &   43 &  36047 &   29 \\
		\textbf{YUV-T} &   877 &  238 &  27146 &  107 &   40 &  36180 &   29 \\
		\midrule
		\text{Runtime} &   885 &  236 &  27238 &  105 &   40 &  35940 &   31 \\
		\bottomrule
	\end{tabular}
	\caption{
		Runtime for each input format and method in milliseconds. The input format (color spaces) doesn't have an influence on the runtime, but addition channels (thermal) may increase the runtime for some algorithms. The last row is the mean runtime of an algorithm.
	}
	\label{runtime}
\end{table}

\begin{table}
	\centering
	\setlength{\tabcolsep}{4.3pt}
	\begin{tabular}{lrrrrrrr}
		\toprule
		{} &  \textbf{CBAD} &  \textbf{GMM} &  \textbf{LOF} &  \textbf{PCA} &  \textbf{RXG} &  \textbf{RXL} &  \textbf{RXM} \\
		\midrule
		\textbf{HLS} &  0.047 &  0.029 &  0.037 &  0.028 &  0.029 &  0.093 &  0.044 \\
		\textbf{HSV} &  0.050 &  0.033 &  0.040 &  0.028 &  0.033 &  0.096 &  0.048 \\
		\textbf{LAB} &  0.029 &  0.028 &  0.038 &  0.023 &  0.028 &  0.077 &  0.031 \\
		\textbf{LUV} &  0.032 &  0.032 &  0.038 &  0.024 &  0.032 &  0.098 &  0.034 \\
		\textbf{RGB} &  0.029 &  0.027 &  0.036 &  0.026 &  0.027 &  0.084 &  0.030 \\
		\textbf{XYZ} &  0.027 &  0.027 &  0.039 &  0.027 &  0.027 &  0.088 &  0.027 \\
		\textbf{YUV} &  0.029 &  0.026 &  0.038 &  0.023 &  0.026 &  0.077 &  0.032 \\
		\midrule
		\text{AUPRC} &  0.035 &  0.029 &  0.038 &  0.025 &  0.029 &  0.088 &  0.035 \\
		\bottomrule \\
	\end{tabular}
	\setlength{\tabcolsep}{4.3pt}
	\begin{tabular}{lrrrrrrr}
		\toprule
		{} &  \textbf{CBAD} &  \textbf{GMM} &  \textbf{LOF} &  \textbf{PCA} &  \textbf{RXG} &  \textbf{RXL} &  \textbf{RXM} \\
		\midrule
		\textbf{HLS-T} &  0.081 &  0.061 &  0.098 &  0.043 &  0.062 &  0.218 &  0.099 \\
		\textbf{HSV-T} &  0.093 &  0.076 &  0.099 &  0.053 &  0.077 &  0.223 &  0.108 \\
		\textbf{LAB-T} &  0.083 &  0.064 &  0.084 &  0.052 &  0.065 &  0.196 &  0.047 \\
		\textbf{LUV-T} &  0.076 &  0.056 &  0.083 &  0.047 &  0.056 &  0.186 &  0.052 \\
		\textbf{RGB-T} &  0.084 &  0.057 &  0.092 &  0.032 &  0.057 &  0.191 &  0.055 \\
		\textbf{XYZ-T} &  0.082 &  0.057 &  0.083 &  0.032 &  0.058 &  0.196 &  0.037 \\
		\textbf{YUV-T} &  0.078 &  0.055 &  0.088 &  0.042 &  0.056 &  0.190 &  0.050 \\
		\midrule
		\text{AUPRC} &  0.082 &  0.061 &  0.090 &  0.043 &  0.062 &  0.200 &  0.064 \\
		\bottomrule
	\end{tabular}
	\caption{
		Area under the precision-recall curve (AUPRC) values for each color space and color anomaly detection method. The scores are obtained from integral images and are averaged over forest landscapes.
		The last row is the mean AUPRC value over all color spaces.
	}
	\label{auprc.forest}
\end{table}

\begin{table}
	\centering
	\setlength{\tabcolsep}{4.3pt}
	\begin{tabular}{lrrrrrrr}
		\toprule
		{} &  \textbf{CBAD} &  \textbf{GMM} &  \textbf{LOF} &  \textbf{PCA} &  \textbf{RXG} &  \textbf{RXL} &  \textbf{RXM} \\
		\midrule
		\textbf{HLS} &  0.219 &  0.255 &  0.022 &  0.239 &  0.256 &  0.215 &  0.275 \\
		\textbf{HSV} &  0.211 &  0.256 &  0.044 &  0.246 &  0.258 &  0.217 &  0.291 \\
		\textbf{LAB} &  0.243 &  0.242 &  0.021 &  0.276 &  0.241 &  0.144 &  0.189 \\
		\textbf{LUV} &  0.234 &  0.230 &  0.020 &  0.229 &  0.228 &  0.136 &  0.185 \\
		\textbf{RGB} &  0.261 &  0.257 &  0.078 &  0.094 &  0.256 &  0.153 &  0.175 \\
		\textbf{XYZ} &  0.254 &  0.254 &  0.019 &  0.078 &  0.253 &  0.155 &  0.144 \\
		\textbf{YUV} &  0.257 &  0.256 &  0.019 &  0.249 &  0.255 &  0.153 &  0.162 \\
		\midrule
		\text{AUPRC} &  0.240 &  0.250 &  0.032 &  0.202 &  0.249 &  0.168 &  0.203 \\
		\bottomrule \\
	\end{tabular}
	\setlength{\tabcolsep}{4.3pt}
	\begin{tabular}{lrrrrrrr}
		\toprule
		{} &  \textbf{CBAD} &  \textbf{GMM} &  \textbf{LOF} &  \textbf{PCA} &  \textbf{RXG} &  \textbf{RXL} &  \textbf{RXM} \\
		\midrule
		\textbf{HLS-T} &  0.258 &  0.293 &  0.077 &  0.292 &  0.294 &  0.268 &  0.296 \\
		\textbf{HSV-T} &  0.259 &  0.303 &  0.082 &  0.302 &  0.304 &  0.268 &  0.316 \\
		\textbf{LAB-T} &  0.305 &  0.309 &  0.082 &  0.306 &  0.310 &  0.222 &  0.250 \\
		\textbf{LUV-T} &  0.295 &  0.302 &  0.082 &  0.267 &  0.300 &  0.212 &  0.249 \\
		\textbf{RGB-T} &  0.312 &  0.312 &  0.086 &  0.187 &  0.312 &  0.225 &  0.260 \\
		\textbf{XYZ-T} &  0.308 &  0.310 &  0.123 &  0.169 &  0.310 &  0.229 &  0.219 \\
		\textbf{YUV-T} &  0.314 &  0.312 &  0.075 &  0.297 &  0.312 &  0.225 &  0.250 \\
		\midrule
		\text{AUPRC} &  0.293 &  0.306 &  0.087 &  0.260 &  0.306 &  0.236 &  0.263 \\
		\bottomrule
	\end{tabular}
	\caption{
		Area under the precision-recall curve (AUPRC) values for each color space and anomaly detection method. The scores are obtained from integral images and are averaged over open landscapes.
		The last row is the mean AUPRC value over all color spaces.
	}
	\label{auprc.open}
\end{table}

\section{Discussion}\label{discussion}
The AUPRC results in Fig. \ref{area} show that all color anomaly detection methods benefit from additional thermal information, but especially in combination with the forest landscapes.

\begin{figure}[H]
	\begin{subfigure}{0.5\linewidth}
		\centering
		\includegraphics[width=0.85\linewidth]{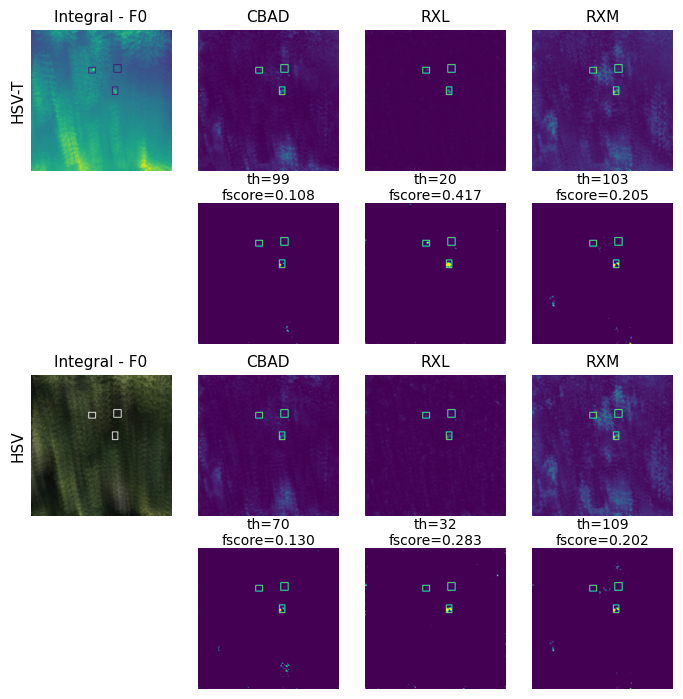}
		\label{scores.F0}
	\end{subfigure}
	\begin{subfigure}{0.5\linewidth}
		\centering
		\includegraphics[width=0.85\linewidth]{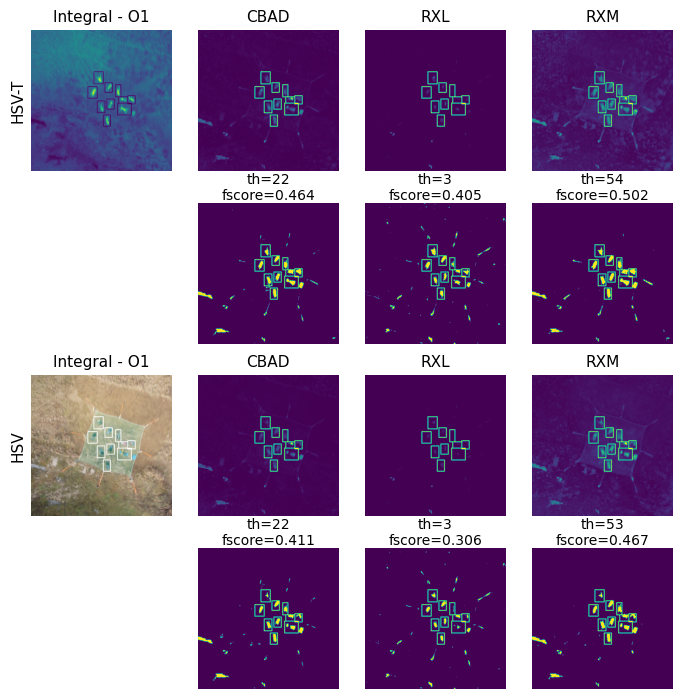}
		\label{scores.O1}
	\end{subfigure}
	\begin{subfigure}{0.5\linewidth}
		\centering
		\includegraphics[width=0.85\linewidth]{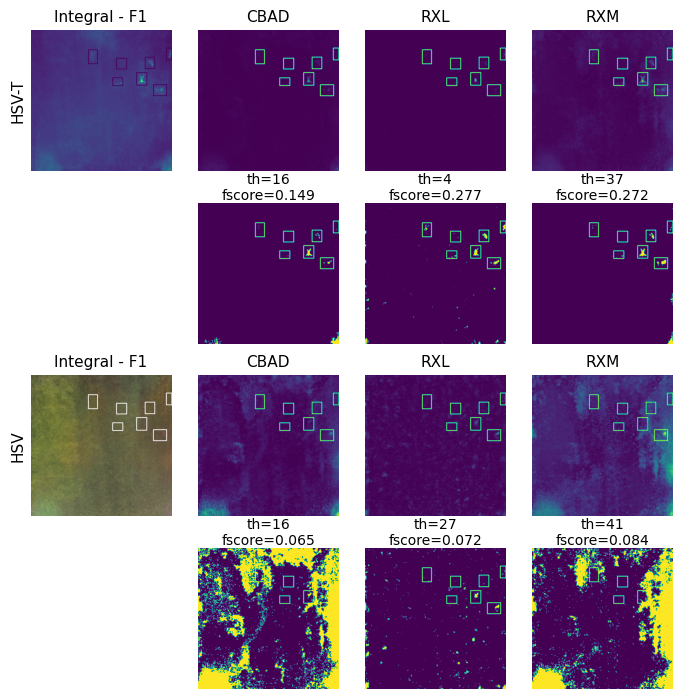}
		\label{scores.F1}
	\end{subfigure}
	\begin{subfigure}{0.5\linewidth}
		\centering
		\includegraphics[width=0.85\linewidth]{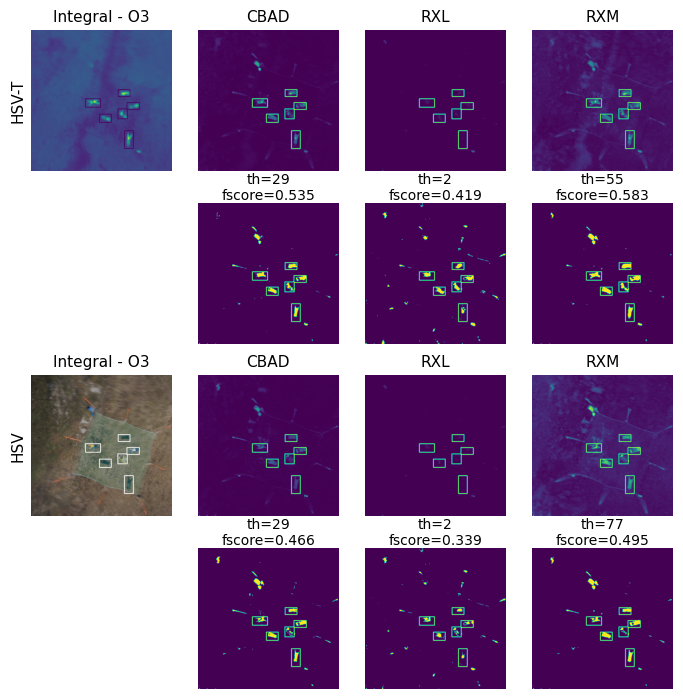}
		\label{scores.O3}
	\end{subfigure}
	\begin{subfigure}{0.5\linewidth}
		\centering
		\includegraphics[width=0.85\linewidth]{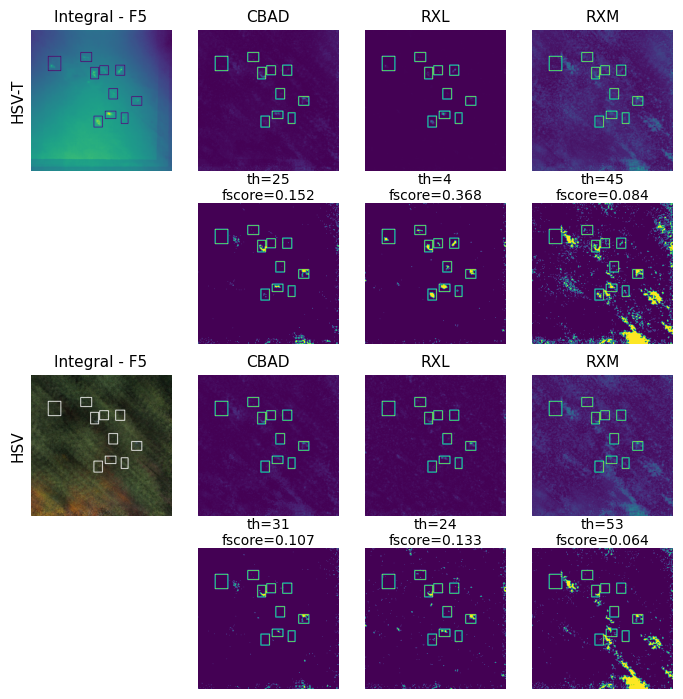}
		\label{Forest Landscape - scores.F5}
	\end{subfigure}
	\begin{subfigure}{0.5\linewidth}
		\centering
		\includegraphics[width=0.85\linewidth]{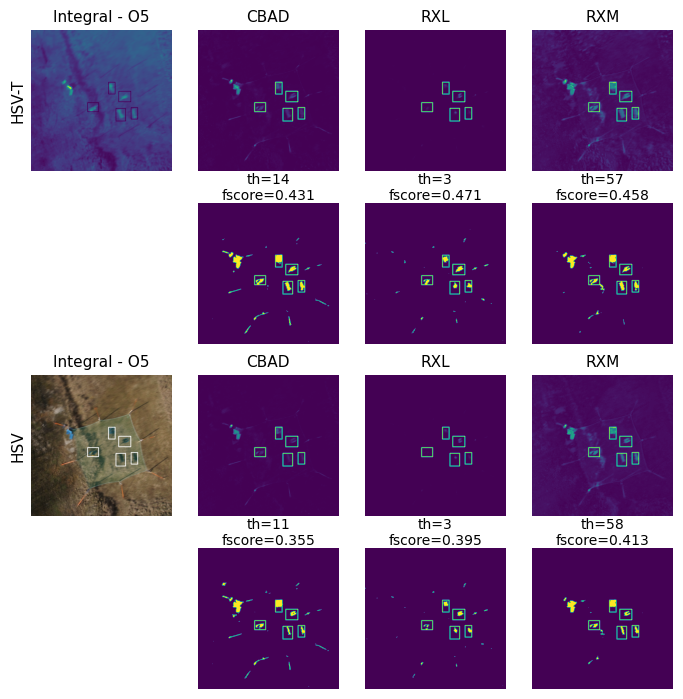}
		\label{scores.O5}
	\end{subfigure}
	\caption{
		Color anomaly detection scores for forest (left) and open (right) landscapes, comparing the overall good performing HSV(-T) inputs.
		The first rows shows anomaly scores of the best (RXL) algorithm without considering runtime and the best (RXM) and second best (CBAD) method when considering runtime.
		The second rows are the anomaly scores after thresholding.
	}
	\label{scores}
\end{figure}

In challenging environments, where the distribution of colors has a much higher variance (e.g., F1 in Fig. \ref{scores}, due to bright sunlight), the additional thermal information improves results significantly.
If the temperature difference between targets and the surrounding is large enough, the thermal spectral band may add spatial information (e.g., distinct clusters of persons), which is beneficial for methods that calculate results based on locality properties (e.g. RXL, LOF).

In forest-like environments, the RXL anomaly detector performs best regardless of the input color space. This  could be explained by the specific characteristics of an integral image.
In case of occlusion, the integration process produces  highly blurred images caused by defocused occluders (forest canopy) above the ground, which results in a much more uniformly distributed background.
Since target pixels on the ground stay in focus, anomaly detection methods like RXL, which calculate background statistics on a smaller window around the target, are benefiting from the uniform distributed (local) background.
The same is true for LOF, where the local density in the blurred background regions is much higher than the local density in the focused target region, resulting in overall better outlier detection rates.
Since most objects in open landscapes are located near the focal plane (i.e., at nearly the same altitude above the ground), there is no out-of-focus effect caused by the integration process.
Thus, these methods do not produce similarly good results for the open landscapes. 

For the forest landscapes, the HSV(-T) and HSL(-T) color spaces consistently give better results than others.
The color spaces HSV (hue, saturation, value) and HSL (hue, saturation, lightness) are both based on cylindrical color space geometries and differ mainly in there last dimension (brightness / lightness).
The first two dimensions (hue, saturation) can be considered more important when distinguishing colors, as the last dimension only describes the value (brightness) or lightness of a color.
We assume that the more uniform background resulting from the integration process, also has a positive effect on the distance metric calculations when those two color spaces are used, especially if the background mainly consists of a very similar color tone.
This is again more pronounced for the forest landscapes than for the open landscapes.

Although the AUPRC results obtained from RXL and LOF are best for the forest landscapes, the high runtime indicate that these methods are impractical for real-time applications.
A trade-off must be made between good anomaly detection results and fast runtime, therefore we  consider the top performing methods that provide reliable results within milliseconds further.

\begin{figure}[H]
	\centering
	\includegraphics[width=\linewidth]{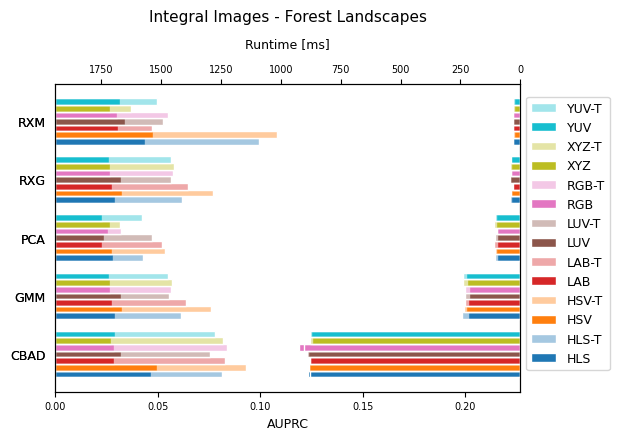}
	\includegraphics[width=\linewidth]{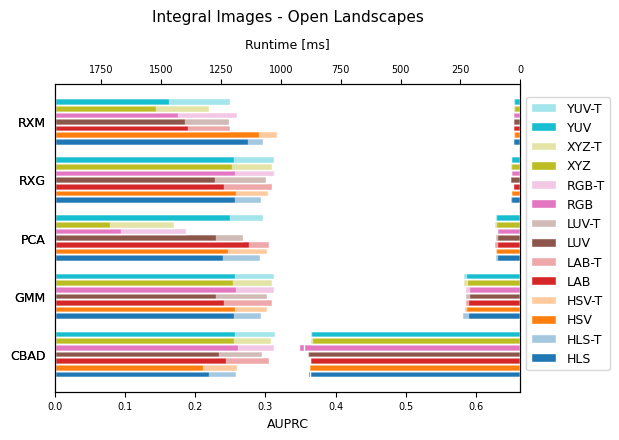}
	\caption{
		Performance in AUPRC (left bars) vs. runtime in $ms$ (right bars): Color anomaly detection methods that produce results in less than a second. Reed Xiaoli Local (RXL) and Local Outlier Factor (LOF) performed well, but needed more than 20 seconds and are therefore not practicable for applications with real-time demands. 
	}
	\label{auprc.runtime}
\end{figure}

Based on the AUPRC and runtime results shown in Fig. \ref{auprc.runtime} one could suggest that the RXM method may be used.
The AUPRC results combined with HSV-T are the best among methods that run under one second, regardless of the landscape.
Since this methods doesn't require a-priory settings to be chosen (only the final thresholding value) and the runtime is one of the fastest, it would be well suited for usage in forests and open landscapes.
The second best algorithm based on the AUPRC values would be CBAD, with the disadvantage that it requires a hyper-parameter setting and doesn't generalize well for open landscapes.

\section{Conclusions}\label{conclusion}
In this article we have shown that the performance of unsupervised color anomaly detection methods applied on multispectral integral images can be further improved by an additional thermal channel.
Each of the evaluated methods performs significant better when thermal information is utilized in addition, regardless of the landscape (forest or open).
Another finding is that even without the additional thermal band, the choice of input color space (for the visible channels) already has an influence on the results.
Color spaces like HSV and HLS can outperform the widely used RGB color space, especially in forest-like landscapes.
These findings might guard decisions on the choice of color anomaly detection method, input format, and applied spectral band, depending on individual use cases. Occlusion cause by vegetation, such as forests, remains challenging for many of them.    

\vspace{6pt}

\authorcontributions{Conceptualization, O.B. and F.S.; methodology, F.S.; software, F.S. and I.K.; validation, F.S., I.K. and O.B.; formal analysis, F.S.; investigation, F.S.; resources, I.K.; data curation, I.K.; writing---original draft preparation, F.S. and O.B.; writing---review and editing, F.S. and O.B.; visualization, F.S.; supervision, O.B.; project administration, O.B.; funding acquisition, O.B. All authors have read and agreed to the published version of the manuscript.}

\funding{This research was funded by the Austrian Science Fund (FWF) under grant number P32185-NBL, and by the State of Upper Austria and the Austrian Federal Ministry of Education, Science and Research via the LIT–Linz Institute of Technology under grant number LIT-2019-8-SEE114.}

\dataavailability{The data and sourcode used in experiments can be downloaded from https://doi.org/10.5281/zenodo.3894773 and https://github.com/JKU-ICG/AOS/.} 

\begin{adjustwidth}{-\extralength}{0cm}
\reftitle{References}
\bibliography{References}
\end{adjustwidth}

\end{document}